\documentclass[11 pt,a4paper]{article}
\usepackage[utf8]{inputenc}
\usepackage{amsmath}
\usepackage{amsfonts}
\usepackage{amssymb}
\usepackage{pgfplots}
\usepackage{tikz}
\usepgfplotslibrary{fillbetween}
\usetikzlibrary{patterns}
\usepackage{indentfirst}
\usepackage{subcaption}
\usetikzlibrary{arrows.meta}
\usepackage{mathtools}
\usepackage{changepage}
\usepackage{amsmath}
\usepackage{lscape}
\newtheorem{thm}{Theorem}
\pgfplotsset{compat=1.15}
\usepackage{multirow}
\usepackage{glossaries}
 \makeglossaries

\author{Dr. Jacques Balayla MD, MPH\footnote{To whom correspondence should be addressed: Dr. Jacques Balayla MD, MPH. Quilligan Scholar. e-mail: jacques.balayla@mcgill.ca. Department of Obstetrics and Gynaecology. McGill University, Montreal, Quebec, Canada}}

\title{Information Threshold,  Bayesian Inference and Decision-Making}

\date{}

\begin{document}

\maketitle  

\begin{abstract}
We define the information threshold as the point of maximum curvature in the prior vs.  posterior Bayesian curve,  both of which are described as a function of the true positive and negative rates of the classification system in question.  The nature of the threshold is such that for sufficiently adequate binary classification systems,  retrieving excess information beyond the threshold does not significantly alter the reliability of our classification assessment.  We hereby introduce the ``marital status thought experiment" to illustrate this idea and report a previously undefined mathematical relationship between the Bayesian prior and posterior,  which may have significant philosophical and epistemological implications in decision theory.  Where the prior probability is a scalar between 0 and 1 given by $\phi$ and the posterior is  a scalar between 0 and 1 given by $\rho$,  then at the information threshold,  $\phi_e$:
\begin{equation}
\phi_e + \rho_e = 1
\end{equation}
Otherwise stated,  given some degree of prior belief,  we may assert its persuasiveness when sufficient quality evidence yields a posterior so that their combined sum equals 1.  Retrieving further evidence beyond this point does not significantly improve the posterior probability,  and may serve as a benchmark for confidence in decision-making. 
\end{abstract} 

\newpage
\section{Glossary of Bayesian Terminology}
Bayes' Theorem has been dubbed as the ``equation of knowledge" \cite{lipovetsky2021equation} given its ubiquitousness in practically all fields of scientific inquiry and its ability to model statistical problems that are frequently encountered in nature.  It isn't therefore surprising that different fields use different terminology to refer to the same concepts evoked in Bayesian theory \cite{van2021bayesian}.  While the following list isn't exhaustive,  it will help the reader navigate this manuscript with ease.
\\
\

\begin{table}[h!]
\centering
\begin{tabular}{|lcll|}
\hline
\multicolumn{4}{|l|}{\textbf{Table 1. Glossary of Bayesian Terminology}} \\ \hline
\multicolumn{1}{|l|}{\textbf{Parameter}} & \multicolumn{1}{l|}{\textbf{Notation}} & \multicolumn{1}{l|}{\textbf{Definition}} & \textbf{Alternative} \\ \hline
\multicolumn{1}{|l|}{\begin{tabular}[c]{@{}l@{}}True Positive \\ Rate\end{tabular}} & \multicolumn{1}{c|}{a} & \multicolumn{1}{l|}{\begin{tabular}[c]{@{}l@{}}Proportion of correct predictions \\ in predictions of positive class. \\ It is defined as the number of true \\ positives over all predicted \\ positive data points.\end{tabular}} & \begin{tabular}[c]{@{}l@{}}Sensitivity, \\ Recall, \\ Hit Rate\end{tabular} \\ \hline
\multicolumn{1}{|l|}{\begin{tabular}[c]{@{}l@{}}True Negative \\ Rate\end{tabular}} & \multicolumn{1}{c|}{b} & \multicolumn{1}{l|}{\begin{tabular}[c]{@{}l@{}}Proportion of correct predictions \\ in predictions of negative class. \\ It is defined as the number of true \\ negatives over all predicted \\ negative data points.\end{tabular}} & \begin{tabular}[c]{@{}l@{}}Specificity, \\ Selectivity\end{tabular} \\ \hline
\multicolumn{1}{|l|}{Prior} & \multicolumn{1}{c|}{$\phi$, P(H)} & \multicolumn{1}{l|}{\begin{tabular}[c]{@{}l@{}}A probability as assessed before \\ making reference to certain \\ relevant observations.\end{tabular}} & \begin{tabular}[c]{@{}l@{}}Pre-test \\ probability\end{tabular} \\ \hline
\multicolumn{1}{|l|}{Posterior} & \multicolumn{1}{c|}{$\rho(\phi)$} & \multicolumn{1}{l|}{\begin{tabular}[c]{@{}l@{}}The statistical probability that \\ a hypothesis is true calculated \\ in the light of relevant \\ observations.\end{tabular}} & \begin{tabular}[c]{@{}l@{}}Post-test \\ probability,\\Positive\\ predictive\\ value,\\ Precision\end{tabular} \\ \hline
\multicolumn{1}{|l|}{\begin{tabular}[c]{@{}l@{}}Likelihood \\ Function\end{tabular}} & \multicolumn{1}{c|}{LR} & \multicolumn{1}{l|}{\begin{tabular}[c]{@{}l@{}}The likelihood that a given \\ outcome would be expected in a \\ positive class compared to \\ the likelihood that it would be \\ expected in a negative class.\end{tabular}} & \begin{tabular}[c]{@{}l@{}}Likelihood \\ ratio\end{tabular} \\ \hline
\multicolumn{1}{|l|}{Evidence} & \multicolumn{1}{c|}{P(E)} & \multicolumn{1}{l|}{\begin{tabular}[c]{@{}l@{}}The probability of an event \\ irrespective of the outcome of \\ another variable.\end{tabular}} & \begin{tabular}[c]{@{}l@{}}Marginal \\ Likelihood\end{tabular} \\ \hline
\end{tabular}
\end{table}

\newpage
\section{Epistemology and the Philosophy of Knowledge}
A proposition is considered to have objective truth when its truth conditions are met without bias caused by a sentient subject \cite{kaur2015ethical}. The truth of a belief or statement is entirely a matter of how things are with its object, and has nothing to do with the state of its subject – the person holding the beliefs about the nature of the object in question \cite{kaur2015ethical}.   However,  how we come to learn about an objective truth - what is known as epistemology,  can be - paradoxically  - a more subjective process \cite{leeds1974subjective}.  Epistemology refers to the philosophical analysis of the nature of knowledge and the conditions required for a belief to constitute knowledge,  such as truth and justification \cite{audi2010epistemology}.  It should come as no surprise therefore that said subjective process is what lies at the core of the differing interpretations of the natural world between beings.  Ultimately,  the degree of belief about a claim will constitute our personal truth,  which, when properly informed and sufficient information has been retrieved,  attempts to match the standard, objective truth.
\section{Bayesianism as a Means to Knowledge}
‘Bayesian epistemology’ became an epistemological movement in the 20th century, though its two main features can be traced back to the eponymous Reverend Thomas Bayes (c. 1701–61) \cite{talbott2001bayesian}. Those two features are: (1) the introduction of a formal apparatus for inductive logic; (2) the introduction of a pragmatic self-defeat test for epistemic rationality as a way of extending the justification of the laws of deductive logic to include a justification for the laws of inductive logic \cite{lee2017preface}.  The formal apparatus itself has two main elements: the use of the laws of probability as coherence constraints on rational degrees of belief (or degrees of confidence) and the introduction of a rule of probabilistic inference, a rule or principle of conditionality \cite{talbott2001bayesian}.
\section{Terminology used in Bayesian Inference}
Given its wide range of applications in a vast number of areas of inquiry,  Bayesian theory employs terminology that may differ between fields.  For the purposes of this research,  we'll refer to $probability$ as the likelihood that an event occurs or that a proposition is true \cite{ramsey2016truth}.  The prior probability reflects the subjective belief about the likelihood of an event $prior$ to new data being collected \cite{lavine1991sensitivity}.  The prior probability of an event will be revised as new data or information becomes available, to produce a more accurate measure of a potential outcome \cite{monahan1992proper}.  That revised probability becomes the $posterior$ probability and is calculated using Bayes' theorem.  As such,  the prior and posterior probabilities are proportional by means of a likelihood term \cite{efron2013bayes}.  
\begin{center}
Posterior $\propto$ Likelihood * Prior
\end{center}
\section{Bayes’ Theorem}
As described,  Bayes’ Theorem provides the mathematical tool that allows for the updating of our beliefs about a hypothesis $H$ in light of some new evidence $E$ \cite{rouder2018teaching}.  Mathematically speaking, the equation translates to the conditional probability of an event or hypothesis  $H$ given the presence of an event or evidence $E$ \cite{garbolino2002evaluation}.   As per Bayes' Theorem, the above relationship is equal to the probability of H given E,  multiplied by the ratio of independent probabilities of event H to event E \cite{westbury2010bayes}.  Simply stated, the equation is written as follows:

\begin{large}
\begin{equation}
P(H|E) = \frac{P(E|H) P(H)}{P(E)}
\end{equation}
\end{large} 

Where H, E are the hypothesis and evidence, respectively,  $P(H|E)$ is the posterior (probability of H given E is true),  P(H) is the prior probability or working hypothesis,  $P(E|H)$ is the likelihood constant (probability of E given H is true),  and P(E) is the marginalization constant, also known as the normalizing constant, or evidence \cite{balayla2020prevalence}. 
\\
\

If we use E +/- as either the presence or absence of evidence,  and denote H +/- as the veracity or falsehood of our hypothesis,  then we can use Bayes' theorem to calculate the posterior probability by asking the following question: given the presence of some evidence $E$, what is the probability that our initial hypothesis $H$ is correct \cite{balayla2020prevalence}?

\begin{large}
\begin{equation}
P(H+|E+)= \frac{P(E+|H+)P(H+)}{P(E+|H+)P(H+)+P(E+|H-)P(H-)}
\end{equation}
\end{large} 

Now,  how do we gather the evidence $E$ to update our beliefs? We collect data and input it into a classification system,  which then categorizes the data into a number of groups depending on the value assigned to each data point \cite{hernandez2008bayes}.  Notably,  these can be classified into groups which represent the positive/normal condition and the others which represent the negative/aberrant condition \cite{mackenzie2014bayesian}.  However,  the data collected itself may be properly classified or not.  As such,  we may have positive or negative data points that are either true or false (Table 2)\cite{balayla2021prevalence}.
\\
\
\begin{table}[h!]
\begin{tabular}{|cc|cc|}
\hline
\multicolumn{2}{|c|}{\multirow{2}{*}{\textbf{Table 2. 2x2 Table}}}             & \multicolumn{2}{c|}{\textbf{Truth}}                                \\ \cline{3-4} 
\multicolumn{2}{|c|}{}                                                         & \multicolumn{1}{c|}{\textbf{Positive (P)}} & \textbf{Negative (N)} \\ \hline
\multicolumn{1}{|c|}{\multirow{2}{*}{\textbf{Assessment}}} & \textbf{Positive} & \multicolumn{1}{c|}{True Positive (TP)}    & False Positive (FP)   \\ \cline{2-4} 
\multicolumn{1}{|c|}{}                                     & \textbf{Negative} & \multicolumn{1}{c|}{False Negative (FN)}   & True negative (TN)    \\ \hline
\end{tabular}
\end{table}

Since by the law of total probability the probability of the hypothesis being false is equal to the complement of it being true (and vice-versa),  and the false positive rate is equal to the complement of the true negative rate (and vice-versa)\cite{balayla2020prevalence},  Bayes' theorem yields the posterior probability, $\rho(\phi)$, as per equation 3, in the following manner:

\begin{large}
\begin{equation}
\rho(\phi) = \frac{a\phi}{ a\phi+(1-b)(1-\phi)}
\end{equation}
\end{large} 
\\
\
where, a = true positive rate (TP/P), b = true negative rate (TN/N),  and $\phi$ = prior probability. We have thus shown that the posterior probability, $\rho$, is a function of the prior $\phi$.  For a set true positive and negative rate,  the greater the confidence in the prior belief,  the greater $\rho(\phi)$ and vice-versa \cite{balayla2021formalism}.
\\
\

\begin{center}

\begin{tikzpicture}
 
	\begin{axis}[
    axis lines = left,
    xlabel = $\phi$,
	ylabel = {$\rho(\phi)$},    
     ymin=0, ymax=1,
    legend pos = south east,
     ymajorgrids=true,
     xmajorgrids=true,
    grid style=dashed,
    width=8cm,
    height=8cm,
     ]
	\addplot [
	domain= 0:1,
	color= blue,
	]
	{(0.95*x)/((0.95*x+(1-0.99)*(1-x))};
	\end{axis} 
\end{tikzpicture}

\textbf{Figure 1.} Prior vs. posterior curve as a function of some binary classification system with some inherent true positive and true negative rate.
\end{center}
Note that the proportionality between the prior and posterior probability is curvilinear in all but a special case \cite{balayla2022bayesian}.  As such, an increase in the prior probability does not necessarily confer a similar increase in the posterior and vice-versa.  Otherwise stated,  the likelihood term relaying the prior to the posterior is not a linear function \cite{balayla2020prevalence}.
\newpage
\subsection{Binary Classification Systems}
A binary classification system (Figure 2) is an assessment tool with two class labels.  In most binary classification problems, one class represents the normal condition and the other represents the aberrant condition \cite{bhamare2016feasibility}.  Given that probabilistic laws are unitary, we can classify anything as either $X$ or $not$ $X$ such that they're collectively exhaustive and mutually exclusive \cite{lee2018mutually}:

\begin{large}
\begin{equation}
X + \neg X = 1
\end{equation}
\end{large}
 	
In other words,  the set containing all possible elements contains a subset X and a subset  $\neg X$ that are mutually exclusive - though in some exceptional cases,  there may be overlap between both groups or one of the groups may be empty.  If we search to identify an element X amongst all possible elements,  then what must remain is the subset of the elements not belonging to X.  To use a simple example,  while we could describe an apple in terms of its shape or color or taste,  one equally valid way to identify an apple would be to say that it is not a banana.  Repeating this task for all fruits in a set would yield two distinct groups, namely, bananas in one group and all other fruits in the other.  The accuracy of binary classification systems is defined as the proportion of correct predictions - both positive and negative - made by a classification model or computational algorithm \cite{chicco2020advantages}.  A value between 0 (no accuracy) and 1 (perfect accuracy),  the accuracy of a classification model is dependent on several factors, notably: the classification rule or algorithm used (what gets classified as positive or negative), the intrinsic characteristics of the tool used to do the classification (what are the inherent true positive and true negative rates of the classification tool),  and the degree of confidence on the prior probability - since for a given likelihood,  higher priors yield higher posteriors \cite{balayla2022bayesian}.
\\
\
\begin{figure}[h!]
\textbf{Figure 2}. BCS's relay prior to posterior probabilities
\centering
\includegraphics[scale=0.34]{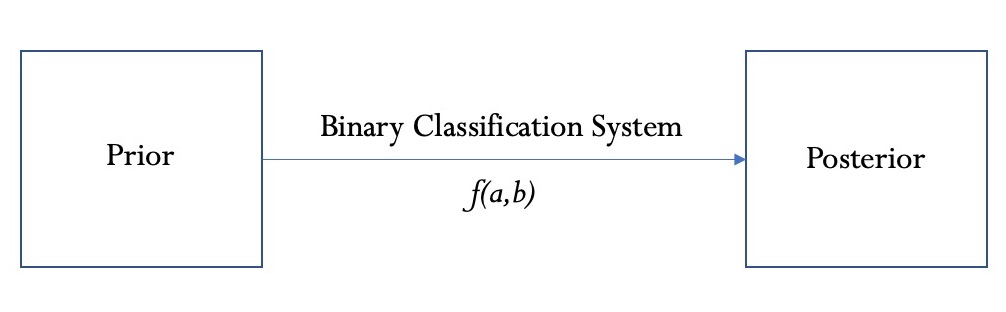}
\end{figure}
\newpage
\subsection{Multiclass Classification Systems}
Since not all elements of a set come in binary form,  we may need a multiclass classification system to organise them \cite{engchuan2015pathway}.  A multiclass classification system classifies the elements of a set into 3 or more categories.  The techniques developed based on reducing the multi-class problem into multiple binary problems can also be called problem transformation heuristic techniques \cite{xu2011extended}.  These can be categorized into ``one vs. rest" and ``one vs. one".  For the purposes of this topic,  we'll focus on the ``one vs. rest" transformation \cite{yu2014comparative}.  
Let S be the following finite set:
\begin{center}
$S$ = $\lbrace X,Y,Z \rbrace$,  where X $\neq$ Y $\neq$ Z
\end{center}
Then,  we have X,Y,Z as proper subsets of S:
\begin{center}
$X \subsetneq S$,  $Y \subsetneq S$, $Z \subsetneq S$
\end{center}
Since intersections of subsets are associative - X$\cap$(Y$\cap$Z) = (X$\cap$Y)$\cap$Z = $\lbrace \rbrace$: 
\begin{center}
$\lbrace Y,Z \rbrace = W= \neg X $
\end{center}
Then,  
\begin{center}
$S$ = $\lbrace X, W \rbrace$ = $\lbrace X, \neg X \rbrace$
\end{center}

Hence whereas originally,  S had 3 mutually exclusive elements,  we perform a ``one vs.  rest" transformation to end with 2 elements in S as a function of X.  While information is indeed lost in subsets Y and Z by grouping them into W,  this transformation allows us to obtain a binary classification from a multiclass classification system where X is maintained in a ``one vs. rest" fashion.  Rendering the classification from a multiclass to a binary form simplifies the Bayesian approach \cite{rocha2013multiclass}.

\section{Properties of the $\rho(\phi)$ Curve}
In order to determine some of the properties of the prior vs. posterior curve,  $\rho(\phi)$, we can study its geometry  \cite{balayla2020prevalence}.  In particular,  the study of the graph's curvature yields important information about how priors impact the posterior in a given binary classification system \cite{balayla2020prevalence}.  Let us first define the curvature $\kappa$ of the function by means of the radius of curvature  $R$ at any given point on the line.  In order to calculate the radius of curvature of the $\rho(\phi)$ graph at any given point M, we consider a circle with radius $R$, which is perpendicular to the tangent line of the function at that point. We consider an adjacent point increment by d$\phi$ and draw another tangent line to this point N, which we join to the center of the circle with radius $R$. As such, an arc of length dS is formed, which in turn creates an angle $\varphi$ between M and N. These variables see the following properties:
\begin{large}
\begin{equation}
tan(\varphi) = \frac{d\rho}{d\phi}
\end{equation}
\end{large}

\begin{large}
\begin{equation}
dS = R{d\varphi} = \sqrt{1 + (\frac{d\rho}{d\phi})^2}d\phi
\end{equation}
\end{large}
\\
\

From equation (6),  the differential equation follows:

\begin{large}
\begin{equation}
\frac{d}{d\phi}tan(\varphi) = \frac{d}{d\phi}(\frac{d\rho}{d\phi}) = \frac{d^2\rho}{d\phi^2}
\end{equation}
\end{large}

From the trigonometric identity $1 + tan^2(\varphi) = sec^2(\varphi)$, it follows that:
\begin{large}
\begin{equation}
 sec^2(\varphi)\frac{d\varphi}{d\phi} = \frac{d^2\rho}{d\phi^2}
\end{equation}
\end{large}

Therefore,

\begin{large}
\begin{equation}
(1 + tan^2(\varphi))\frac{d\varphi}{d\phi} = \frac{d^2\rho}{d\phi^2}
\end{equation}
\end{large}

Since $tan(\varphi) = d\rho/d\phi$, equation (10) becomes:

\begin{large}
\begin{equation}
(1 + (\frac{d\rho}{d\phi})^2)\frac{d\varphi}{d\phi} = \frac{d^2\rho}{d\phi^2}
\end{equation}
\end{large}
\\
\

Isolating $d\varphi/d\phi$, we obtain:

\begin{large}
\begin{equation}
\frac{d\varphi}{d\phi} = \frac{\frac{d^2\rho}{d\phi^2}}{(1 + (\frac{d\rho}{d\phi})^2)}
\end{equation}
\end{large}

Using equation (7) this relationship then becomes:

\begin{large}
\begin{equation}
R\frac{\frac{d^2\rho}{d\phi^2}}{(1 + (\frac{d\rho}{d\phi})^2)} = \sqrt{1 + (\frac{d\rho}{d\phi})^2}
\end{equation}
\end{large}
\newpage
Finally, isolating the radius of curvature $R$:

\begin{large}
\begin{equation}
R= \frac{[1 + (\frac{d\rho}{d\phi})^2]^\frac{3}{2}}{|\frac{d^2\rho}{d\phi^2}|}
\end{equation}
\end{large}

The radius of curvature $R$ is inversely proportional to $\kappa$ such that:
\begin{large}
\begin{equation}
R = \frac{1}{\kappa} \Rightarrow \kappa = \frac{|\frac{d^2\rho}{d\phi^2}|}{[1 + (\frac{d\rho}{d\phi})^2]^\frac{3}{2}}
\end{equation}
\end{large}

Now that we know what the curvature function $\kappa$ is, we can determine where the curvature of $\rho(\phi)$ falls at a maximum. Practically speaking, this represents the point of sharpest change in $\frac{d\rho}{d\phi}$, known as the extrema. 
In order to do so, we find the derivative of the $\kappa$ function and determine its roots:
\begin{large}
\begin{equation}
\frac{d\kappa}{d\phi}=0 \hookrightarrow\lbrace\phi_e,\rho_e\rbrace 
\end{equation}
\end{large}

The above equation yields the value of $\phi$ where the maximum curvature $\kappa$ and thus a minimum radius of curvature $R$ exist. We define this point as the point of local extrema [$\phi_e,\rho_e$] of the $\rho(\phi)$ function. On the other hand, the inflection point [$\phi_i,\rho_i$] is a point on a curve at which the sign of the curvature (i.e., the concavity) changes. 
\\
\ 

The points of local extrema are distinguishable from the inflection point only in that the curvature function's second order-derivative equals 0:

\begin{large}
\begin{equation}
\frac{d^2\kappa}{d\phi^2}=0 \hookrightarrow\lbrace\phi_i,\rho_i\rbrace 
\end{equation}
\end{large}

However, as we described previously, given the proportionality between $\phi$ and $\rho$ all screening curves retain their concavity/convexity throughout the domain [0,1] as a function of the true positive and negative rates, and thus no inflection points are observed in these curves. Conversely, the point of local extrema $\phi_e,\rho_e$ tells us where the sharpest turn, or change, in posteriors as a function of the prior occurs. The first order derivative of $\rho(\phi)$ is given by:

\begin{equation}
\rho'(\phi)=\frac{a\left(-b+1\right)}{\left(a\phi+\left(1-b\right)\left(1-\phi)\right)\right)^2}
\end{equation}
\newpage
By equating equation (15) to 0 and using equation (18) to find its roots, we re-arrange the terms and the above expression simplifies to:

\begin{large}
\begin{equation}
1 = -\frac{a^{2}\left(-b+1\right)^{2}+\left(a\phi+\left(-b+1\right)\left(-\phi+1\right)\right)^{4}}{2\left(a\phi+\left(-b+1\right)\left(-\phi+1\right)\right)^{4}}
\end{equation}
\end{large}

\begin{large}
\begin{equation}
\left(a\phi+\left(-b+1\right)\left(-\phi+1\right)\right)^{4} = -a^{2}\left(-b+1\right)^{2}
\end{equation}
\end{large}

Taking the fourth root of both sides, we obtain:

\begin{large}
\begin{equation}
\left(a\phi+b\phi-b-\phi+1\right) = \pm\sqrt{a\left(-b+1\right)}
\end{equation}
\end{large}

Expanding and isolating $\phi$ while taking the positive value of the root so that the value obtained may fall inside the domain of the function, we obtain:
\begin{large}
\begin{equation}
\phi_e = \frac{\sqrt{a\left(-b+1\right)}+b-1}{(a+b-1)}
\end{equation}
\end{large}

This is the priors value where the point of local extrema $\phi_e$ of $\rho(\phi)$ is found. We denote this value of $\phi$ as the \textit{prevalence} or \textit{information} threshold.  Note the inverse relationship between $\phi_e$ and  Youden's $J$ statistic \cite{youden1950index},  given that $J$ = a+b-1.

\begin{large}
\begin{equation}
\phi_e \sim \frac{1}{J}
\end{equation}
\end{large}

Using radical conjugates, we can further simplify the threshold equation into its most basic form - without the need for the Youden's $J$ statistic.  Let c = 1-b, the complement of the true negative rate, otherwise known as the fall-out or false positive rate (FPR). The $\phi_e$ equation thus becomes:

\begin{large}
\begin{equation}
\phi_e = \frac{\sqrt{ac}-c}{a-c}
\end{equation}
\end{large}

Multiplying by its radical conjugate, we obtain:

\begin{large}
\begin{equation}
\phi_e = \frac{\sqrt{ac}-c}{a-c} \left[\frac{\sqrt{ac}+c}{\sqrt{ac}+c}\right]
\end{equation}
\end{large}

The square difference in the numerator yields:

\begin{large}
\begin{equation}
\phi_e = \frac{ac-c^2}{a\sqrt{ac}+ac-c\sqrt{ac}-c^2} 
\end{equation}
\end{large}

Factoring out c, and knowing that $\frac{\sqrt{x}}{x}$ is equal to $\frac{1}{\sqrt{x}}$we obtain:

\begin{large}
\begin{equation}
\phi_e = \frac{a-c}{{\frac{a}{c}\sqrt{ac}}+a-\sqrt{ac}-c} = \frac{a-c}{{a\frac{\sqrt{a}}{\sqrt{c}}-\sqrt{ac}}+a-c}
\end{equation}
\end{large}

Factoring out $\sqrt{ac}$ in the denominator's first terms leads to:

\begin{large}
\begin{equation}
\phi_e =  \frac{a-c}{{\sqrt{ac}(\frac{a}{c}-1)}+a-c}
\end{equation}
\end{large}

Finally, replacing 1 by c/c and factoring out the ensuing a-c term, we obtain:
\begin{large}
\begin{equation}
\phi_e =  \left[\frac{a-c}{a-c}\right]\frac{1}{\frac{\sqrt{a}}{\sqrt{c}}+1}
\end{equation}
\end{large}

And thus, replacing c by 1-b the simplified version of the equation follows:
\begin{large}
\begin{equation}
\phi_e = \frac{1}{{\sqrt\frac{a}{1-b}}+1}=\frac{\sqrt{1-b}}{\sqrt{a}+\sqrt{1-b}}
\end{equation}
\end{large}

Using this threshold as a prior value, we can calculate the corresponding posterior by plotting $\phi_e$ into the posterior equation to retrieve [$\rho(\phi_e)$,$\phi_e$]. In so doing we obtain:

\begin{large}
\begin{equation}
\rho(\phi_e)=\sqrt{\frac{a}{1-b}}\left[\frac{\sqrt{1-b}}{\sqrt{a}+\sqrt{1-b}}\right]
\end{equation}
\end{large}

Interestingly, the above expression leads to the well known formulation of the posterior as  a function of priors and the likelihood  term,  in some fields known as the positive likelihood ratio (LR+) \cite{kent1982robust}, defined as the true positive rate over the compliment of the true negative rate.

\begin{large}
\begin{equation}
\rho(\phi_e)=\phi_e\sqrt{\frac{a}{1-b}} = \phi_e  \sqrt{LR+}
\end{equation}
\end{large}
\

We have previously shown that relative to a perfect accuracy of 1, the information threshold bounds the accuracy of binary classification systems \cite{balayla2021prevalence}. Indeed, where computational time is a limiting resource, attaining the information threshold in binary classification systems may be sufficient to yield levels of accuracy comparable to that under maximum priors \cite{balayla2021prevalence}. 
\section{Graphical representation of the Bayesian curve}
\begin{center}
\begin{tikzpicture}
 
	\begin{axis}[
    axis lines = left,
    xlabel = $\phi$,
	ylabel = {$\rho(\phi)$},    
     ymin=0, ymax=1,
    legend pos = south east,
     ymajorgrids=true,
     xmajorgrids=true,
    grid style=dashed,
    width=8cm,
    height=8cm,
     ]
	\addplot [
	domain= 0:1,
	color= blue,
	]
	{(0.95*x)/((0.95*x+(1-0.99)*(1-x))};
	\end{axis} 
		 \draw [dashed, red, thick] (0.56,0) -- (0.56,5.76);
		
\end{tikzpicture}
\end{center}
\begin{center}
\textbf{Figure 3.} Prior vs. posterior curve and the information threshold (red).
\end{center}
\begin{center}
\begin{tikzpicture}
 
	\begin{axis}[
    axis lines = left,
    xlabel = $\phi$,
	ylabel = {$\rho(\phi)$},    
     ymin=0, ymax=1,
    legend pos = south east,
     ymajorgrids=true,
     xmajorgrids=true,
    grid style=dashed,
    width=8cm,
    height=8cm,
    style={ultra thick}
     ]
	\addplot [
	domain= 0:0.093,
	color= blue,
	]
	{(0.95*x)/((0.95*x+(1-0.99)*(1-x))};
	
		\addplot [
	domain= 0.093:1,
	color= green,
	]
	{(0.95*x)/((0.95*x+(1-0.99)*(1-x))};
	\end{axis} 
		 \draw [dashed, red, thick] (0.56,0) -- (0.56,5.76);
		
\end{tikzpicture}
\end{center}
\begin{center}
\textbf{Figure 4.} The information threshold divides the Bayesian curve into two partitions, a more vertical section below and a horizontal one beyond,  where increasing the priors does not significantly improve the posterior. The greater the curvature, the more pronounced the two partitions.  Otherwise stated, the greater the true positive and negative rates, the greater the curvature and the greater the area under the curve \cite{balayla2020prevalence}.
\end{center}
\newpage
\subsection{Interesting Properties of the Information Threshold}
We have previously defined the $information$ threshold as the inflection point of maximum curvature in the prior vs.  posterior curve.  Below this point,  the rate of change of the posterior drops at a differential pace relative to the prior.  In practical terms,  below this point,  type I errors -  the mistaken rejection of a true null hypothesis \cite{rothman2010curbing},  what is colloquially known as ``false positives" - increase.   For illustrative purposes,  Figure 5 demonstrates several prior vs.  posterior curves, each with individual true positive and true negative rate parameters,  chosen indiscriminately.

\begin{center}

\begin{tikzpicture}
 
	\begin{axis}[
    axis lines = left,
    xlabel = $\phi$,
	ylabel = {$\rho(\phi)$},    
     ymin=0, ymax=1,
    legend pos = south east,
     ymajorgrids=true,
     xmajorgrids=true,
    grid style=dashed,
    width=6cm,
    height=6cm,
     ]
     \addplot [
	domain= 0:1,
	color= blue,
	]
	{(0.95*x)/((0.95*x+(1-0.99)*(1-x))};
	 \addplot [
	domain= 0:1,
	color= orange,
	]
	{(0.85*x)/((0.85*x+(1-0.95)*(1-x))};
    \addplot [
	domain= 0:1,
	color= black,
	dashed,
	]
	{(1-x)};	
	\addplot [
	domain= 0:1,
	color= red,
	]
	{(0.75*x)/((0.75*x+(1-0.85)*(1-x))}; 
	\addplot [
	domain= 0:1,
	color= gray,
	]
	{(0.5*x)/((0.5*x+(1-0.5)*(1-x))};
	\addplot [
	domain= 0:1,
	color= black,
	]
	{(0.2*x)/((0.2*x+(1-0.4)*(1-x))};
	\addplot [
	domain= 0:1,
	color= magenta,
	]
	{(0.1*x)/((0.1*x+(1-0.1)*(1-x))};
	\addplot [
	domain= 0:1,
	color= brown	,
	]
	{(0.02*x)/((0.02*x+(1-0.02)*(1-x))};
	\end{axis}
\end{tikzpicture}

\textbf{Figure 5.  Bayesian curves with different true positive and negative rates}
\end{center}
Note that regardless of the true positive and negative rate combination,  all threshold points of maximum curvature lie along the y = 1-x line, which means that at the information threshold the sum of the prior and the posterior equal 1:
\begin{equation}
\boxed{\rho_e + \phi_e = 1}
\end{equation}
\begin{table}[h!]
\centering
\begin{tabular}{|ccccc|}
\hline
\multicolumn{5}{|l|}{\textbf{Table 3. Examples of Bayesian Curves}}                                                                                         \\ \hline
\multicolumn{1}{|c|}{\textbf{TPR}} & \multicolumn{1}{c|}{\textbf{TNR}} & \multicolumn{1}{c|}{\textbf{$\phi_e$}} & \multicolumn{1}{c|}{\textbf{$\rho(\phi_e)$}} & \textbf{$\Sigma_{\rho_e + \phi_e}$} \\ \hline
\multicolumn{1}{|c|}{0.95}         & \multicolumn{1}{c|}{0.99}         & \multicolumn{1}{c|}{0.093}       & \multicolumn{1}{c|}{0.917}              & 1            \\ \hline
\multicolumn{1}{|c|}{0.85}         & \multicolumn{1}{c|}{0.95}         & \multicolumn{1}{c|}{0.195}       & \multicolumn{1}{c|}{0.815}              & 1            \\ \hline
\multicolumn{1}{|c|}{0.75}         & \multicolumn{1}{c|}{0.85}         & \multicolumn{1}{c|}{0.309}       & \multicolumn{1}{c|}{0.691}              & 1            \\ \hline
\multicolumn{1}{|c|}{0.50}         & \multicolumn{1}{c|}{0.50}         & \multicolumn{1}{c|}{0.500}       & \multicolumn{1}{c|}{0.500}              & 1            \\ \hline
\multicolumn{1}{|c|}{0.20}         & \multicolumn{1}{c|}{0.40}         & \multicolumn{1}{c|}{0.633}       & \multicolumn{1}{c|}{0.367}              & 1            \\ \hline
\multicolumn{1}{|c|}{0.10}         & \multicolumn{1}{c|}{0.10}         & \multicolumn{1}{c|}{0.750}       & \multicolumn{1}{c|}{0.250}              & 1            \\ \hline
\multicolumn{1}{|c|}{0.02}         & \multicolumn{1}{c|}{0.02}         & \multicolumn{1}{c|}{0.875}       & \multicolumn{1}{c|}{0.125}              & 1            \\ \hline
\multicolumn{5}{|c|}{\begin{tabular}[c]{@{}c@{}}TPR, TNR = true positive, negative rate \end{tabular}}                     \\ \hline
\end{tabular}
\end{table}

\newpage
\begin{thm}
Let $\rho(\phi)$ be the equation for the Bayesian posterior probability as a function of the prior $\phi$ and some likelihood term $H$, which is itself a function of the binary classification system's true positive and negative rates.  Then,  there exists an information threshold - $\phi_e$ - so that, for sufficiently adequate binary classification systems, increasing the prior probability does not significantly increase the posterior probability beyond this point.  At this prior level,  $\rho(\phi_e)$ + $\phi_e$ = 1
\end{thm}
\subsection{Proof of $\rho_e + \phi_e = 1$}
In order to prove that $\rho_e + \phi_e = 1$ at the information threshold,  we simply add the posterior equation to the prior parameter and then input the information threshold variable as the prior term to obtain:
\begin{large}
\begin{equation}
\rho(\phi) + \phi = \frac{a\phi}{ a\phi+(1-b)(1-\phi)} + \phi = \frac{a\phi^2+\phi-\phi^2-b\phi+\phi^2b+a\phi}{a\phi+\left(1-b\right)\left(1-\phi\right)}
\end{equation}
\end{large} 
Isolating $\phi$, we obtain:
\begin{large}
\begin{equation}
=\frac{\phi\left(a\phi+1-\phi-b+\phi b+a\right)}{\phi\left(a+\frac{1}{\phi}\:-1-\frac{b}{\phi}+b\right)}\: = \frac{\left(a\phi+1-\phi-b+\phi b+a\right)}{\left(a+\frac{1}{\phi}\:-1-\frac{b}{\phi}+b\right)}\: 
\end{equation}
\end{large} 
Inputting the information threshold parameter,  $\phi_e$, we obtain:
\begin{large}
\begin{equation}
=\frac{\left(a\frac{1}{\sqrt{\frac{a}{1-b}}+1}+1-\frac{1}{\sqrt{\frac{a}{1-b}}+1}-b+\frac{1}{\sqrt{\frac{a}{1-b}}+1}b+a\right)}{\left(a+\frac{1}{\frac{1}{\sqrt{\frac{a}{1-b}}+1}}\:-1-\frac{b}{\frac{1}{\sqrt{\frac{a}{1-b}}+1}}+b\right)}\:
\end{equation}
\end{large} 
Simplifying the numerator and denominator:
\begin{large}
\begin{equation}
=\frac{\frac{2a+\sqrt{\frac{a}{1-b}}-b\sqrt{\frac{a}{1-b}}+a\sqrt{\frac{a}{1-b}}}{\sqrt{\frac{a}{1-b}}+1}}{a+\sqrt{\frac{a}{1-b}}-b\sqrt{\frac{a}{1-b}}}
\end{equation}
\end{large} 
Applying the fraction rule:
\begin{large}
\begin{equation}
=\frac{2a+\sqrt{\frac{a}{1-b}}-b\sqrt{\frac{a}{1-b}}+a\sqrt{\frac{a}{1-b}}}{\left(\sqrt{\frac{a}{1-b}}+1\right)\left(a+\sqrt{\frac{a}{1-b}}-b\sqrt{\frac{a}{1-b}}\right)}= \frac{2a+ \sqrt{\frac{a}{1-b}}(a-b+1)}{a+(\frac{a}{1-b})(1-b)+\sqrt{\frac{a}{1-b}}(a-b+1)} 
\end{equation}
\end{large} 
As the numerator and denominator cancel, we obtain:
\begin{large}
\begin{equation}
\rho_e + \phi_e = 1
\end{equation}
\end{large} 
\boxed{$Q.E.D.$}
\newpage
\subsection{The Marital Status Thought Experiment}
We can more simply illustrate the principle of updating priors and information threshold with a descriptive illustration.  Intuitively,  we know that in order to recognize and identify the presence of an object,  a pattern,  and even abstract constructs like illness,  the nature of reality or the veracity of some claim,  requires some input data or information.  We likewise have all experienced situations whereby the amount of information necessary to reach a definitive conclusion is often a fraction of all the information available.  We define ``definitive" in this context less as an indicator of an indisputable or incontrovertible fact but more as the level of persuasion required to deem our conclusion personally satisfactory.  In other words,  this is the point where we're likely to have retrieved sufficient information to make informed decisions and reach conclusions based on our prior beliefs.  Retrieving further evidence beyond this point does not significantly improve the posterior probability,  and may therefore serve as a benchmark for confidence in decision-making.  It is nevertheless critical to specify that this process has no claim on the objective truth.  We may very well be wrong in our assessment's conclusion.  But the information threshold may represent a saturation point where we're satisfied with the degree of confidence in our prior belief and the evidence we have obtained to make a decision or conclude our assessment.  In these scenarios, our intelligence and our senses act as a classification system subject to Bayesian rules too. 
\\
\

To better illustrate this notion - we developed a thought experiment - henceforth known as the ``marital status thought experiment".  Say that you see a person walking down the street: what are the chances that person is married? Say now they’re holding the hand of a person of the opposite sex.  What are the chances that person is married? Say now that there is a child in between them.  What are the chances that person is married? Say now they’re each wearing a wedding band. What are the chances that person is married? Say now you check their ID’s and notice they share an address. 
What are the chances that person is married? Evidently,  each scenario updates and increases your confidence in the belief the couple is married.   We could in theory take this argument ad infinitum,  always finding additional clues that may suggest the individual in question is married.  We know that beyond a certain arbitrary point - adding further information does not significantly improve nor changes our predictions. That point represents an information  threshold that is specific to the binary classification system in question - the human being's ability to classify each scenario or data point  whereby evidence in favor of marriage gets assigned a positive value and evidence against gets assigned a negative value.  The argument brought forth by this theory is that a persuasive conclusion can be reached when the evidence updates our priors to the point where $\rho_e + \phi_e = 1$.  
\subsection{What constitutes a sufficiently adequate BCS?}
The theorem above suggests that for sufficiently adequate binary classification systems, the sum of the prior and posterior represents a point of sufficient information to make decisions, since, as per the theory, improving the priors does not significantly increase our posterior beyond this point.  The important question thus becomes, what constitutes a ``sufficiently adequate binary classification system"? Theoretically, the best possible binary classification system has true positive and negative rates equal to 1. In essence,  this is a perfect classifier.  Unfortunately,  given natural variation between individuals and systems,  this is in fact quite difficult to attain in real life.  We need therefore a system that is as close to the standard as possible.  Power et al. \cite{power2013principles} propose a rule of thumb whereby the sum of the true positive and negative rates add to at least 1.5 to consider a binary classifier as adequate.  We have previously defined the variable $\varepsilon$ as the sum of true positive and negative rates \cite{balayla2020prevalence}, where:
\begin{large}
\begin{equation}
\varepsilon = a + b  \rightarrow [\varepsilon\in {\rm I\!R} | 0 < \varepsilon < 2] \rightarrow J = \varepsilon - 1
\end{equation}
\end{large}  
 Others suggest that at least 90$\%$ true positive and negative rates ought to be considered satisfactory.  In truth, the answer isn't as straightforward since the binary classifiers are more sensitive to changes in the true negative rate than the true positive rate.  This fact is beyond the scope of this manuscript,  but we can nevertheless use the area under the curve (AUC) of $\rho(\phi)$ to aid in the answer to this question.  Let us define:
 
 \begin{equation}
 \int{\rho(\phi)d\phi} > \lambda
 \end{equation}
 where $\lambda$ is some standard area under the curve for some true positive and negative rate.  Expanding the indefinite integral, we obtain:
\

\begin{equation}
 \int{\rho(\phi)d\phi}=\frac{a\left(\phi\left(a+b-1\right)+\left(b-1\right)\ln\left(\phi\left(a+b-1\right)-b+1\right)\right)}{\left(a+b-1\right)^{2}}
\end{equation}
Introducing the domain [0,1] removes $\phi$ as a variable from the equation:
\begin{equation}
 \int_{0}^{1}{\rho(\phi)d\phi}=\frac{a\ \left(a+(b-1)\ln\left(a\right)\right)}{\left(a+b-1\right)^{2}}-\frac{a\left(\left(1-b\right)+\left(b-1\right)\ln\left(1-b\right)\right)}{\left(a+b-1\right)^{2}}
\end{equation}

\begin{equation}
=\frac{a}{(a+b-1)^{2}}\cdot  [(a+(b-1)\ln\left(a\right) -(1-b)+\left(b-1\right)\ln\left(1-b\right)]
\end{equation}
\newpage
Given Youden's J statistic = $(a+b-1)$:
\begin{equation}
 =\frac{a^2+a(b-1)\ln(a)-a(1-b)-a(b-1)\ln(1-b)]}{J^2} > \lambda
\end{equation}

By the logarithmic rules:
\begin{equation}
 =\frac{a^2+a(b-1)[\ln(\frac{a}{1-b})+1]}{J^2} > \lambda
\end{equation}

What $\lambda$ level one chooses is a matter of context and degree of confidence necessary. There's a well-established statistical precedent in Science whereby we're willing to incur a 5$\%$ chance of being incorrect in our assessment.  This is closely related to using the 95$\%$ confidence as a landmark of statistical significance.  If we adopt a similar approach and are willing to incur a 5$\%$ chance of a type-I error relative to a perfect accuracy, then $\lambda$ would be equal to 0.95,  such that, within the domain [0,1],  we obtain:
  \begin{equation}
 \int_{0}^{1}{\rho(\phi)d\phi} > 0.95
 \end{equation}
The question then becomes, what parameters of $a$ and $b$ yield areas greater than $\lambda$? First, as mentioned previously, the system is more sensitive to changes in the true negative rate, $b$.  This comes from the term $\ln(\frac{a}{1-b})$ where greater values of $b$ increases the value of $(\frac{a}{1-b})$.  In order to avoid the use of partial differential equations,  we can use brute force methods to deduce the bounds of $a$ and $b$ in each scenario:
\subsubsection{Scenario 1: $b > a$}
In cases of binary classifiers whose true negative rate is greater than its true positive rate, we may consider a 0.99 true negative rate to determine how much loss the true positive rate could afford to maintain an area under the curve such that $\lambda$ $>$ 0.95 (the function is not defined at b = 1):
\

\begin{equation}
 =\frac{a^2+a(0.99-1)[\ln(\frac{a}{1-0.99})+1]}{(a+0.99+1)^2} >0.95\leftrightarrow a \approx 0.667
\end{equation}
\
Thus,
\
\begin{center}
$b>a\Rightarrow \lambda > 0.95 \Leftrightarrow a > 0.66,  \varepsilon \approx 0.99 + 0.66 \approx 1.65$
\end{center}
\

In such cases where $b>a$ and $b$ = 0.99,  an adequate binary classification system would have the sum $\varepsilon$ of the true positive and negative rates add to at least 1.65. This would imply an information threshold of:
\

\begin{equation}
\phi_e = \frac{\sqrt{1-b}}{\sqrt{a}+\sqrt{1-b}} = \frac{\sqrt{1-0.99}}{\sqrt{0.66}+\sqrt{1-0.99}} = 0.109
\end{equation}
\

As per equation (39), $\rho(\phi_e)$ = 0.89, such that 
$\rho(\phi_e)$ + $\phi_e$ $\approx$ 1.  In other words,  the ratio of posterior to prior $\approx$ 9:1 in order to ensure an accuracy over 0.95. 

\subsubsection{Scenario 1: $a > b$}
In cases of binary classifiers whose true positive rate is greater than its true negative rate, we may consider a 0.99 true positive rate to determine how much loss the true negative rate could afford to maintain an area under the curve such that $\lambda$ $>$ 0.95:
\

\begin{equation}
 =\frac{0.99^2+0.99(b-1)[\ln(\frac{0.99}{1-b})+1]}{(0.99+b+1)^2} >0.95\leftrightarrow b \approx 0.982
\end{equation}
\
Thus,
\
\begin{center}
$a>b\Rightarrow \lambda > 0.95 \Leftrightarrow b > 0.982,  \varepsilon \approx 0.99 + 0.98 \approx 1.97$
\end{center}
\

In such cases where $a>b$ and $a$ = 0.99,  an adequate binary classification system would have the sum of the true positive and negative rates add to at least 1.97.  Again, the difference is not that surprising since binary classifiers in this context are much more sensitive to minute changes in the true negative rate. This would imply an information threshold of:

\begin{equation}
\phi_e = \frac{\sqrt{1-b}}{\sqrt{a}+\sqrt{1-b}} = \frac{\sqrt{1-0.98}}{\sqrt{0.99}+\sqrt{1-0.98}} \approx 0.124
\end{equation}
\

As per equation (39), $\rho(\phi_e)$ = 0.87, such that \ 
$\rho(\phi_e)$ + $\phi_e$ $\approx$ 1.  In other words,  the ratio of posterior to prior $\approx$ 9:1 as well.  Table 4 reports the expected minimal bounds on the true positive and negative rates as a function of $\lambda$.
\\
\

\textbf{Table 4. Expected minimal bounds on the true positive and negative rates as a function of $\lambda$}
\\
\

\begin{table}[h!]
\centering
\begin{tabular}{|ccccccc|}
\hline
\multicolumn{1}{|c|}{\textbf{Table 4.}} & \multicolumn{3}{c|}{\textbf{a = 0.99}} & \multicolumn{3}{c|}{\textbf{b = 0.99}} \\ \hline
\multicolumn{1}{|c|}{\textbf{$\lambda$}} & \multicolumn{1}{c|}{\textbf{b}} & \multicolumn{1}{c|}{\textbf{$\phi_e$}} & \multicolumn{1}{c|}{\textbf{$\approx$ Ratio}} & \multicolumn{1}{c|}{\textbf{a}} & \multicolumn{1}{c|}{\textbf{$\phi_e$}} & \textbf{$\approx$ Ratio} \\ \hline
\multicolumn{1}{|c|}{0.95} & \multicolumn{1}{c|}{0.985} & \multicolumn{1}{c|}{0.109} & \multicolumn{1}{c|}{9:1} & \multicolumn{1}{c|}{0.66} & \multicolumn{1}{c|}{0.11} & 9:1 \\ \hline
\multicolumn{1}{|c|}{0.90} & \multicolumn{1}{c|}{0.96} & \multicolumn{1}{c|}{0.16} & \multicolumn{1}{c|}{8.5:1.5} & \multicolumn{1}{c|}{0.25} & \multicolumn{1}{c|}{0.16} & 8.5:1.5 \\ \hline
\multicolumn{1}{|c|}{0.85} & \multicolumn{1}{c|}{0.925} & \multicolumn{1}{c|}{0.21} & \multicolumn{1}{c|}{8:2} & \multicolumn{1}{c|}{0.13} & \multicolumn{1}{c|}{0.21} & 8:2 \\ \hline
\multicolumn{1}{|c|}{0.80} & \multicolumn{1}{c|}{0.87} & \multicolumn{1}{c|}{0.26} & \multicolumn{1}{c|}{7.5:2.5} & \multicolumn{1}{c|}{0.08} & \multicolumn{1}{c|}{0.26} & 7.5:2.5 \\ \hline
\multicolumn{7}{|l|}{*These values are approximated to the second decimal point} \\ \hline
\end{tabular}
\end{table}
\newpage
\subsection{Statistical Parameters of Binary Classification Systems}
Several examples of binary classifications systems exist.  In medical science, screening tests are binary classification systems that categorize individual as sick or not.  In artificial intelligence (AI),  machine learning algorithms serve as classification systems as well,  amongst many others.  Could we extrapolate the statistical parameters of classification systems to human beings? That is, could we assign a value for true positive and negative rates to a particular human being in a specific context, since after all, our senses and the cognitive processes they elicit help us navigate the world by engaging in constant, active classification? A preliminary search of the literature reveals no such parameters.  One potential cause may be that unlike an inorganic,  lifeless classification tool,  humans have cognitive abilities that are subject to bias,  since we would by definition hold both the priors and the mental processes that perform the classification to arrive at the posteriors.  However,  there is no inherent obstacle to consider human beings and their senses as classification tools as well.
Indeed, the posterior is interpreted as a summary of two sources of information:  the subjective beliefs or the information possessed before observing the data - the priors, and the information provided by the data following classification.  Being able to summarize these two sources of information in a single object (the posterior) is one of the main strengths of the Bayesian approach.
\subsection{How are decisions made?}
In its simplest sense, decision-making is the act of choosing between two or more courses of action, often with the intent of maximizing utility and minimising loss or adversity of outcome.  In the wider process of problem-solving, decision-making involves choosing between different possible solutions to a problem \cite{simon1977new}.   This idea evokes a strong parallel to binary and multiclass classification problems and plays a critical role in economic theory,  game theory, risk management, and several other disciplines.  Decisions can be made through either an intuitive or reasoned process, or most often,  a combination of the two \cite{simon1977new}.  Intuition and reasoning are therefore both critical to make decisions.  Intuition refers to the instinctive feelings (rather than conscious reasoning) that some claim is true in the absence of evidence.  It is informed by prior experience and logic as well as by drawing parallels to similar scenarios from which we may deduce some reality.  Such a definition for intuition closely relates to the definition of the prior probability in Bayesian inference: the degree of belief about the truth of a claim as assessed before making reference to certain relevant observations, especially subjectively or on the assumption that all possible outcomes be given the same probability.  Likewise,  conscious reasoning is defined as the part of the decision-making process which uses evidence gathered from observation to reach conclusions. Here too,  this definition closely relates to the definition of the posterior probability: the statistical probability that a hypothesis is true calculated in the light of relevant observations.  It is on this basis that the branch of Bayesian decision theory has flourished.  Given the proportionality between the prior and the posterior,  and given that beyond the information threshold the posterior does not significantly change with increasing priors,  attaining the threshold would imply that when the sum of your subjective intuition and conscious reasoning equal one, decisions may be made, as beyond the prior level, the posterior is unlikely to change.  It is worth noting that we operate under the impression that priors take place $before$ the posterior,  and indeed this is most often the case,  as suggested in the name.  However,  when making the parallel between a Bayesian prior and intuition, we can consider scenarios whereby the intuition is less about the claim and more about the posterior.  For example,  you may be presented with some evidence $E$ regarding some claim you have no previous familiarity with.  Making an intuition about the nature of the posterior itself acts as a prior, and is thus subject to $\rho(\phi_e)$ + $\phi_e$ $\approx$ 1, too.  Indeed, the relationship is commutative. 
\section{Philosophical implications and conclusion}
Intelligence is defined in a myriad of ways.  At its core, however, intelligence reflects the ability to adequately and correctly classify things into their right category so we avoid the pitfalls and dangers that come with Life.  Indeed,  we classify (or categorize) things, people, ideas, feelings in ways that maximize our happiness and thus our survival – this is why intelligence is an important distinctive feature between humans and other species.  However,  as described in this manuscript,  our classification ability is limited by Bayes’ Theorem, which creates an existential dilemma with regards to decision-making since retrieving further information is always, in theory, possible yet may be unnecessary. Likewise,  below the threshold, the reliability of our classification ability may be compromised.  Despite carrying vast implications, the information threshold is a simple metric that determines the  level where a tool’s precision and thus by extension — our understanding of reality— begins to fail.  Developing better observation tools and obtaining sufficient information to stay above the information threshold is necessary to draw more reliable conclusions about our observations in the world.  Otherwise stated,  given some degree of prior belief,  we may assert its persuasiveness when sufficient quality evidence yields a posterior so that their combined sum equals 1.  Retrieving further evidence beyond this point does not significantly improve the posterior probability,  and may serve as a benchmark for confidence in decision-making.

\newpage
\bibliographystyle{unsrt}
\bibliography{references}

\begin{thebibliography}{10}

\bibitem{lipovetsky2021equation}
Stan Lipovetsky.
\newblock The equation of knowledge: From bayes’ rule to a unified philosophy
  of science, 2021.

\bibitem{van2021bayesian}
Rens van~de Schoot, Sarah Depaoli, Ruth King, Bianca Kramer, Kaspar
  M{\"a}rtens, Mahlet~G Tadesse, Marina Vannucci, Andrew Gelman, Duco Veen,
  Joukje Willemsen, et~al.
\newblock Bayesian statistics and modelling.
\newblock {\em Nature Reviews Methods Primers}, 1(1):1--26, 2021.

\bibitem{kaur2015ethical}
Ravneet Kaur.
\newblock Ethical considerations in professional excellence.
\newblock {\em International Journal of Marketing and Technology}, 5(6):80--86,
  2015.

\bibitem{leeds1974subjective}
Anthony Leeds.
\newblock ‘subjective’and ‘objective’in social anthropological
  epistemology.
\newblock In {\em Philosophical foundations of science}, pages 349--361.
  Springer, 1974.

\bibitem{audi2010epistemology}
Robert Audi.
\newblock {\em Epistemology: A contemporary introduction to the theory of
  knowledge}.
\newblock Routledge, 2010.

\bibitem{talbott2001bayesian}
William Talbott.
\newblock Bayesian epistemology.
\newblock 2001.

\bibitem{lee2017preface}
Kevin~P Lee.
\newblock A preface to the philosophy of legal information.
\newblock {\em SMU Sci. \& Tech. L. Rev.}, 20:277, 2017.

\bibitem{ramsey2016truth}
Frank~P Ramsey.
\newblock Truth and probability.
\newblock In {\em Readings in formal epistemology}, pages 21--45. Springer,
  2016.

\bibitem{lavine1991sensitivity}
Michael Lavine.
\newblock Sensitivity in bayesian statistics: the prior and the likelihood.
\newblock {\em Journal of the American Statistical Association},
  86(414):396--399, 1991.

\bibitem{monahan1992proper}
John~F Monahan and Dennis~D Boos.
\newblock Proper likelihoods for bayesian analysis.
\newblock {\em Biometrika}, 79(2):271--278, 1992.

\bibitem{efron2013bayes}
Bradley Efron.
\newblock Bayes' theorem in the 21st century.
\newblock {\em Science}, 340(6137):1177--1178, 2013.

\bibitem{rouder2018teaching}
Jeffrey~N Rouder and Richard~D Morey.
\newblock Teaching bayes’ theorem: Strength of evidence as predictive
  accuracy.
\newblock {\em The American Statistician}, 2018.

\bibitem{garbolino2002evaluation}
Paolo Garbolino and Franco Taroni.
\newblock Evaluation of scientific evidence using bayesian networks.
\newblock {\em Forensic Science International}, 125(2-3):149--155, 2002.

\bibitem{westbury2010bayes}
Chris~F Westbury.
\newblock Bayes’ rule for clinicians: an introduction.
\newblock {\em Frontiers in psychology}, 1:192, 2010.

\bibitem{balayla2020prevalence}
Jacques Balayla.
\newblock Prevalence threshold ($\phi$ e) and the geometry of screening curves.
\newblock {\em Plos one}, 15(10):e0240215, 2020.

\bibitem{hernandez2008bayes}
Daniel Hern{\'a}ndez-Lobato and Jos{\'e}~Miguel Hern{\'a}ndez-Lobato.
\newblock Bayes machines for binary classification.
\newblock {\em Pattern Recognition Letters}, 29(10):1466--1473, 2008.

\bibitem{mackenzie2014bayesian}
Cameron~A MacKenzie, Theodore~B Trafalis, and Kash Barker.
\newblock A bayesian beta kernel model for binary classification and online
  learning problems.
\newblock {\em Statistical Analysis and Data Mining: The ASA Data Science
  Journal}, 7(6):434--449, 2014.

\bibitem{balayla2021prevalence}
Jacques Balayla.
\newblock Prevalence threshold and bounds in the accuracy of binary
  classification systems.
\newblock {\em arXiv preprint arXiv:2112.13289}, 2021.

\bibitem{balayla2021formalism}
Jacques Balayla.
\newblock On the formalism of the screening paradox.
\newblock {\em Plos one}, 16(9):e0256645, 2021.

\bibitem{balayla2022bayesian}
Jacques Balayla.
\newblock Bayesian updating and sequential testing: Overcoming inferential
  limitations of screening tests.
\newblock {\em BMC medical informatics and decision making}, 22(1):1--8, 2022.

\bibitem{bhamare2016feasibility}
Deval Bhamare, Tara Salman, Mohammed Samaka, Aiman Erbad, and Raj Jain.
\newblock Feasibility of supervised machine learning for cloud security.
\newblock In {\em 2016 International Conference on Information Science and
  Security (ICISS)}, pages 1--5. IEEE, 2016.

\bibitem{lee2018mutually}
Chia-Yen Lee and Bo-Syun Chen.
\newblock Mutually-exclusive-and-collectively-exhaustive feature selection
  scheme.
\newblock {\em Applied Soft Computing}, 68:961--971, 2018.

\bibitem{chicco2020advantages}
Davide Chicco and Giuseppe Jurman.
\newblock The advantages of the matthews correlation coefficient (mcc) over f1
  score and accuracy in binary classification evaluation.
\newblock {\em BMC genomics}, 21(1):1--13, 2020.

\bibitem{engchuan2015pathway}
Worrawat Engchuan and Jonathan~H Chan.
\newblock Pathway activity transformation for multi-class classification of
  lung cancer datasets.
\newblock {\em Neurocomputing}, 165:81--89, 2015.

\bibitem{xu2011extended}
Jianhua Xu.
\newblock An extended one-versus-rest support vector machine for multi-label
  classification.
\newblock {\em Neurocomputing}, 74(17):3114--3124, 2011.

\bibitem{yu2014comparative}
Cao Yu, Li~Jinxu, Zhao Fudong, Bian Ran, and Liu Xia.
\newblock Comparative study on face recognition based on svm of one-against-one
  and one-against-rest methods.
\newblock In {\em 2014 8th International Conference on Future Generation
  Communication and Networking}, pages 104--107. IEEE, 2014.

\bibitem{rocha2013multiclass}
Anderson Rocha and Siome~Klein Goldenstein.
\newblock Multiclass from binary: Expanding one-versus-all, one-versus-one and
  ecoc-based approaches.
\newblock {\em IEEE Transactions on Neural Networks and Learning Systems},
  25(2):289--302, 2013.

\bibitem{youden1950index}
William~J Youden.
\newblock Index for rating diagnostic tests.
\newblock {\em Cancer}, 3(1):32--35, 1950.

\bibitem{kent1982robust}
John~T Kent.
\newblock Robust properties of likelihood ratio tests.
\newblock {\em Biometrika}, 69(1):19--27, 1982.

\bibitem{rothman2010curbing}
Kenneth~J Rothman.
\newblock Curbing type i and type ii errors.
\newblock {\em European journal of epidemiology}, 25(4):223--224, 2010.

\bibitem{power2013principles}
Michael Power, Greg Fell, and Michael Wright.
\newblock Principles for high-quality, high-value testing.
\newblock {\em BMJ Evidence-Based Medicine}, 18(1):5--10, 2013.

\bibitem{simon1977new}
H~Simon.
\newblock The new science of management decisionprentice-hall.
\newblock {\em Englewood Cliffs, NJ}, 1977.

\end{thebibliography}

\end{document}